\documentclass[preprint,12pt]{elsarticle}

\usepackage{amssymb}

\usepackage{url}
\usepackage{hyperref}
\usepackage{soul}
\usepackage{listings}
\usepackage{xcolor}
\definecolor{backcolour}{rgb}{0.95,0.95,0.92}
\definecolor{premierEmerald}{HTML}{00b564}
\definecolor{premierEmerald+}{HTML}{006337}
\definecolor{premierPurple}{HTML}{7030A0}
\definecolor{premierPurple-}{HTML}{bc71f5}
\usepackage{tablefootnote}
\usepackage{tikz}
\usetikzlibrary{shapes.geometric}
\usepackage{amsmath}

\journal{Computer Speech \& Language}

\begin{document}

\begin{frontmatter}



\title{The MERIT Dataset: Modelling and Efficiently Rendering Interpretable Transcripts}
\author[1]{Ignacio de Rodrigo\corref{cor1}}
\ead{iderodrigo@comillas.edu}
\cortext[cor1]{Corresponding author}

\author[1]{Alberto Sanchez-Cuadrado}
\ead{ascuadrado@alu.icai.comillas.edu}

\author[1]{Jaime Boal}
\ead{jboal@comillas.edu}

\author[1]{Alvaro J. Lopez-Lopez}
\ead{allopez@comillas.edu}

\affiliation[1]{organization={Institute for Research in Technology, ICAI School of Engineering, Comillas Pontifical University},
            addressline={Calle Santa Cruz de Marcenado, 26}, 
            city={Madrid},
            postcode={28015}, 
            state={Madrid},
            country={Spain}}

\begin{abstract}
This paper introduces the MERIT Dataset, a multimodal (text + image + layout) fully labeled dataset within the context of school reports. Comprising over 400 labels and 33k samples, the MERIT Dataset is a valuable resource for training models in demanding Visually-rich Document Understanding (VrDU) tasks. By its nature (student grade reports), the MERIT Dataset can potentially include biases in a controlled way, making it a valuable tool to benchmark biases induced in Language Models (LLMs). The paper outlines the dataset's generation pipeline and highlights its main features in the textual, visual, layout, and bias domains. To demonstrate the dataset's utility, we present a benchmark with token classification models, showing that the dataset poses a significant challenge even for SOTA models and that these would greatly benefit from including samples from the MERIT Dataset in their pretraining phase.
\end{abstract}




\end{frontmatter}


\section{Introduction}
\label{sec:intro}

Data gathering and synthetic generation are key points to improve AI efficiency, quality, and explainability. Its relevance is sometimes overlooked in both academia and the private sector. Still, several factors (technical and contextual) justify the exploration and exploitation of cheap, accurate, and relevant methods to obtain data.

From a technical perspective, Synthetic Dataset Generation (SDG) involves digitally creating (and sometimes labeling) samples for training Deep Learning (DL) models. SDG techniques are relevant to multiple domains, from image rendering to text generation. SDGs' main challenge is reducing the gap between synthetic datasets and real samples. Conceptually, the examples in the datasets used to train DL models are instances of a multivariate probability distribution, which is the mathematical representation of the reality we aim to model in each case. The success of these DL models relies on two factors: 1) the training process is designed and executed in a way that avoids learning the specific dataset details (noise) associated with the training examples, and 2) the dataset samples to train the models accurately represent the true distribution we want to capture. In other words, it is necessary to represent or capture the essential variation factors to solve the problem. There are primarily two approaches to achieve this: implicitly capturing them in the parameters of a model (a Generative Adversarial Network \cite{goodfellow2020generative}, for instance) or explicitly capturing them by generating synthetic examples in a controlled manner. The first option may be the only alternative for very general or complex problems but poses challenges when generating samples with a high degree of control or specificity. It also often raises numerical challenges in the learning process of the implicit sample generator. The second option is more limited in representing realities with many variation factors but maximizes control over the samples.

On the other hand, we can observe clear dynamics emerging within the context of AI: it has moved from research laboratories to the everyday scene. This has been possible thanks to improvements in model architecture, progress in available hardware for training, and the user-friendly interfaces of Large Language Models (LLMs) and their numerous applications for the general public. A dilemma arises in this constant and rapid improvement scenario: exploit or explore. In this dichotomy, mainstream development (both in the private and academic sectors) has embraced an exploitation stance towards architecture, heavily focused on achieving results. This trend has been even more evident in the case of LLMs, models that scale very successfully and can solve a wide range of tasks (initially in the textual domain and later by introducing the concept of multimodality: text \cite{devlin2018bert}, image \cite{dosovitskiy2020image}, audio\cite{verma2021audio}, or layout \cite{Xu2021}). This strategy has favored the emergence of model families that, within a short period, have exploited architectures by increasing their number of parameters and, thereby, their capabilities.

In this race to scale models, the interest and analysis of training datasets have taken a backseat in some applications. This lack of attention is evident when examining, for example, the datasets used for training some multimodal models. In these cases, the available samples are scanned documents, such as FUNSD \cite{Jaume2019}, XFUND \cite{Xu2022}, CORD \cite{Park2019}, or SROIE \cite{huang2019icdar2019} datasets. This fact implies certain limitations, such as the lack of flexibility in generating the samples (the generation process is highly labor-intensive, making any modifications to the data highly inefficient).

In addition, the established methodology in DL is clear: large institutions with technical, economic, and knowledge resources are the ones capable of developing models from scratch, while the end-user must adapt pre-trained models to their problem domain using techniques like Transfer Learning or Fine Tuning. Therefore, the end-user must have an appropriate and high-quality dataset representing their problem. To cite a few examples, this working method has demonstrated its validity and versatility in models like YOLO \cite{redmon2016you}, based on Convolutional Neural Networks (CNN), or the Trasnformer-based architecture \cite{Vaswani2017} models, with examples like the LayoutLM family \cite{Xu2020} for Visually-rich Document Understanding (VrDU) tasks, or language classification tasks \cite{ali2022hate}.

Furthermore, there are problems where high-quality data are scarce. One of the identified niches is the industrial sector, where, due to data protection policies, it is difficult to find public datasets containing relevant information for such problems. Additionally, the industry's dynamic nature requires end-users to have a fast and agile methodology to adapt models to their working conditions. Synthetic sample generation expands the information that is otherwise impossible to obtain through traditional sample generation methods. In addition, in contrast to conventional sample generation, SDG techniques allow for reducing the human time cost to zero, expanding the available information in classical labeling techniques, and streamlining the study of stimulus-effect explainability of models. Finally, synthetic datasets allow for modeling reality and enable the inclusion of biases in a controlled and scoped manner. Generating these biases facilitates the design of benchmarks in controlled environments to measure model biases and devise firewall policies against potential misuse, with a prominent use case in LLMs \cite{hofmann2024dialect} and its direct applications, for instance, on speech recognition \cite{feng2024towards}.

\begin{figure}
\centering
\includegraphics[width=1\columnwidth]{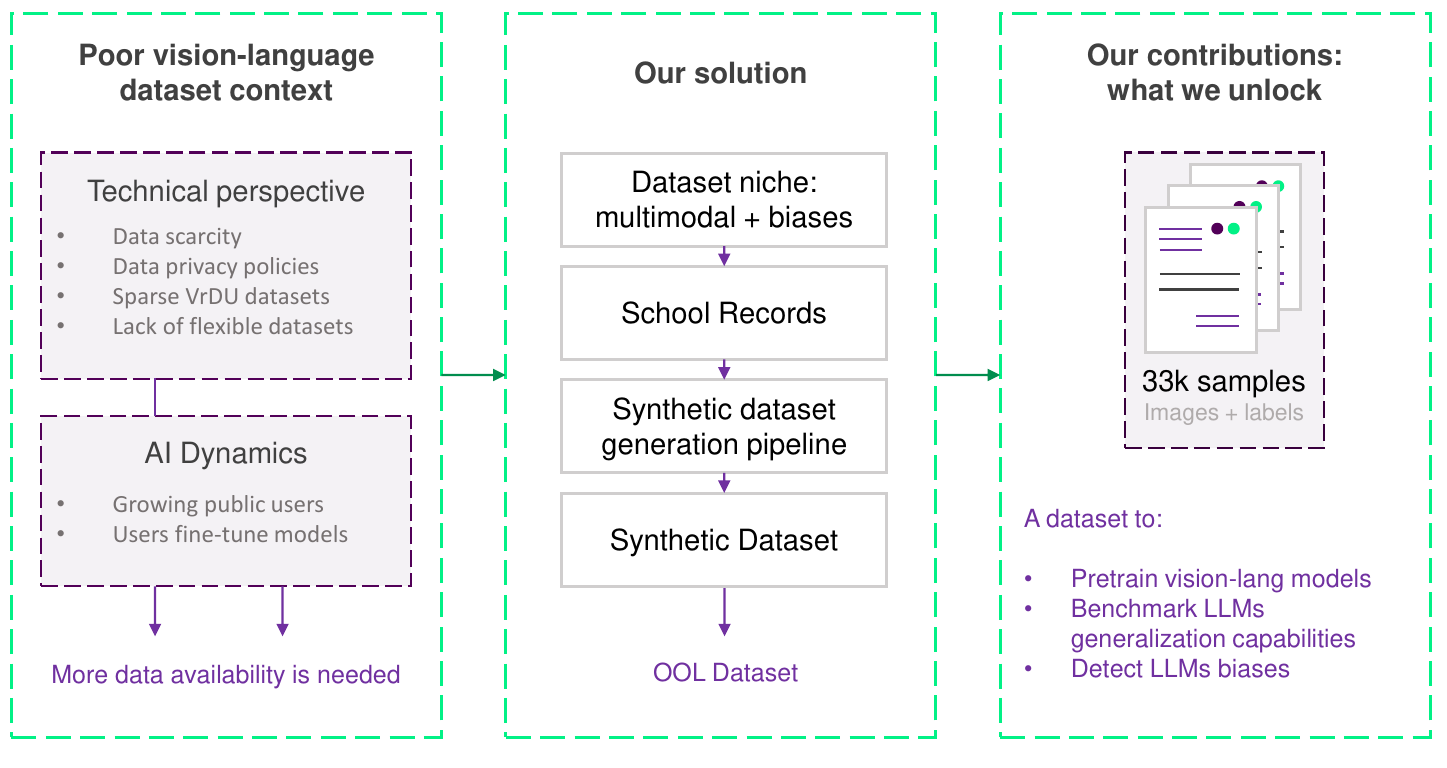}
\caption{Visual abstract: The leftmost block summarizes the AI context, calling for further exploration of synthetic data generation. The center block outlines our approach to generating a detailed dataset and providing a robust pipeline. The rightmost block highlights our contributions and valuable niches for the dataset.}
\label{fig:visualAbstract}
\end{figure}

All these technical reasons (data scarcity, outdated data, or limited flexibility) and the organization of the community (divided into model generators and model users) push for exploring the generation of synthetic datasets for Transformer-based architectures in the context of document scraping or Visually-rich Document Understanding (VrDU). This task has already been tackled with some of the already mentioned datasets (FUNSD \cite{Jaume2019}, CORD \cite{Park2019}, SROIE \cite{huang2019icdar2019}, etc.), enabling SOTA models to achieve excellent metrics \cite{Huang2022}, \cite{kim2021donut}, \cite{tang2023unifying}. However, these models still struggle to reduce generalization errors when applied to more demanding contexts. These contexts often involve a more significant number of classes than those found in FUNSD\cite{Jaume2019} or more complex layouts than those found in CORD\cite{Park2019}. Consequently, an opportunity arises to create a dataset of greater technical complexity. We identify the context of school reports as an ideal niche for elaborating this dataset, given the multitude of labels present (such as subjects and grades, categorized by type and grade level) and the diverse layout formats used to present key information. At last, this context also satisfies the bias-potential requirement: the school reports context also features elements that introduce biases, such as the origin and gender associated with the name on each sample and the grades obtained. Figure \ref{fig:visualAbstract} summarizes the exposed context and relays our approach.

\begin{figure}
\centering
\includegraphics[width=0.75\columnwidth]{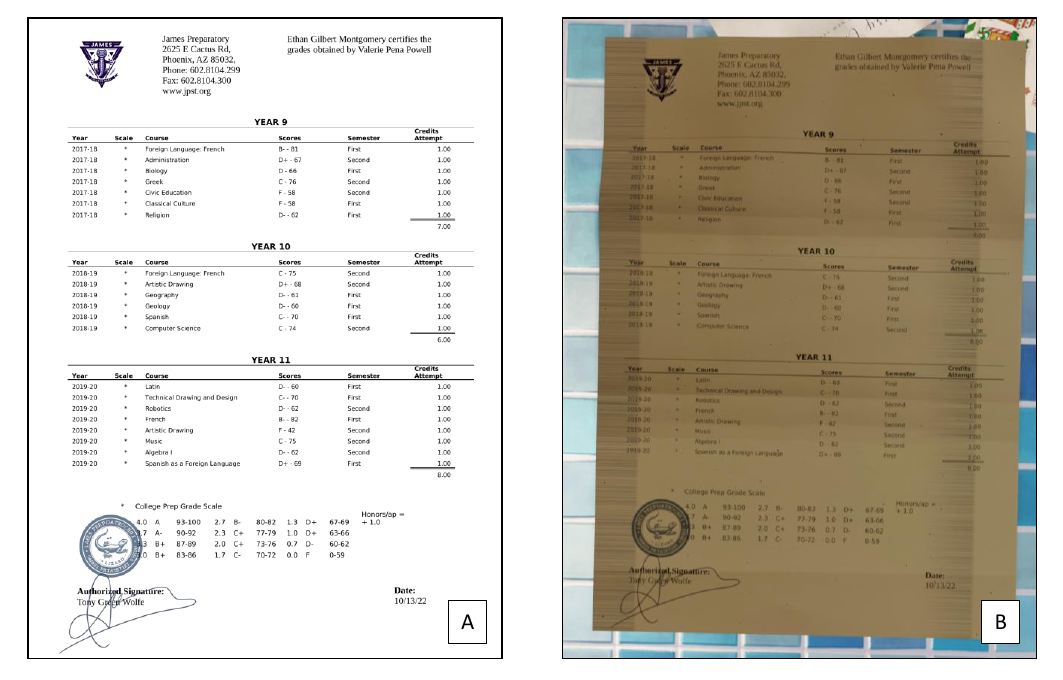}
\caption{Visual styles of the samples. A digital document sample (A) is the output from the first module of the pipeline. A physical document sample (B) is produced by processing sample A through the photorealistic Blender module.}
\label{fig:digitalAndPhysicalSamples}
\end{figure}

This paper introduces the MERIT Dataset and describes its generation pipeline. The MERIT Dataset is a multimodal dataset comprising synthetic digital and photorealistic images labeled within the context of school reports (Figure \ref{fig:digitalAndPhysicalSamples}.A and \ref{fig:digitalAndPhysicalSamples}.B, respectively). It serves as a valuable resource for improving model performance in the Visually-rich Document Understanding (VrDU) task, assessing how multimodal Language Models (LLMs) generalize, and aiding in identifying and mitigating biases within LLMs.

Our main contributions by introducing this dataset and paper are:

\begin{itemize}
    \item Provision of a multimodal (text + image + layout) fully labeled dataset for Visually-rich Document Understanding (VrDU), comprising 33k samples. The dataset is publicly available on \href{https://huggingface.co/datasets/de-Rodrigo/merit}{Hugging Face} \footnote{Dataset on \href{https://huggingface.co/datasets/de-Rodrigo/merit}{Hugging Face}: https://huggingface.co/datasets/de-Rodrigo/merit}.
    \item Presentation of a detailed and comprehensive pipeline for replicating, modifying, or extending the dataset. The code is publicly available on \href{https://github.com/nachoDRT/MERIT-Dataset}{GitHub} \footnote{Code on \href{https://github.com/nachoDRT/MERIT-Dataset}{GitHub}: https://github.com/nachoDRT/MERIT-Dataset}.
    \item Establishment of a benchmark to demonstrate the dataset's effectiveness in training relevant models.
    \item Creation of a synthetic controlled-biased dataset to address data privacy policies and evaluate biases in LLMs.
\end{itemize}

The paper begins by reviewing the related work in Section\ref{sec:relatedWork}. Then, we describe our pipeline to generate our synthetic dataset in Section \ref{sec:datasetGeneration}, describing the samples generation process and the Blender module that modifies them. Section \ref{sec:datasetAnalysis} describes the MERIT Dataset structure and its layout, textual, visual, and ethical features. In Section \ref{sec:experiments}, we benchmark our dataset to prove its suitability to solve a token classification task. Finally, we discuss our contributions and outline our future research based on the MERIT Dataset in Section \ref{sec:discussion}. 

\section{Related work}
\label{sec:relatedWork}

The pursuit of enhanced document understanding systems has been marked by significant strides in the development of sophisticated datasets and the adoption of novel methodologies to interpret complex document structures, i.e., there has been a concerted effort towards improving results by advancing both datasets and models.

There is a wide range of different non-synthetic datasets depending on the task they are designed for. They may include data in various domains, such as text, or combined with images and even layout. Some of them focus on specific tasks such as Named Entity Recognition. One relevant dataset is the NER Dataset \cite{ringland2019nne}, which has finely-grained nested labels to give each word a richer semantic and syntactic context. It is important to note that this implementation elaborates on the Penn Treebank Dataset \cite{marcus1993building} (a dataset limited to the textual domain), and its expansion (nested labels) is carried out through manual human labeling. FUNSD \cite{Jaume2019} is another widely used Dataset for training models in tasks such as Token Classification (TC). FUNSD includes real samples (image domain plus text and layout labels) of scanned documents in English. The structure of these forms varies, and the field of application is diverse. However, the authors point out the cost of including samples from other application fields (since the dataset is not generated from a synthetic pipeline). Other datasets like XFUND \cite{Xu2022} try to solve the language restriction. This dataset includes real document samples in English, Italian, or Japanese (it includes up to seven different languages). In addition, it consists of a more extensive corpus than that offered by FUNSD. Its creation has involved around 1500 hours of human labor, as it is a real dataset with 1393 documents completely hand-tagged. Another domain for labeled document datasets is purchase receipts. Datasets like CORD \cite{Park2019} or SROIE \cite{huang2019icdar2019} (11k and 1k labeled samples, respectively) stand out in this domain. Models trained with these datasets solve text localization or key information extraction tasks. Other datasets, like Publaynet \cite{zhong2019publaynet}, specialize in document layout analysis tasks. This dataset gathers 360k images of digitally born documents focusing on the scientific publications field. This dataset’s limitation is that it comprises un-scanned or photographed documents (so the domain gap might arise when inferring models with real scanned data).
Also, like the rest of the previous cases, its theme is closed and rigid: authors do not offer a flexible mechanism for generating datasets with different typologies or structures. It is also worth mentioning DocVQA \cite{mathew2021docvqa}, a dataset created to train models in the (Visual) Question Answering (QA) task. Building on this foundation, the PDF-VQA \cite{ding2023vqa} and SlideVQA \cite{tanaka2023slidevqa} datasets extend document understanding to encompass multiple pages and incorporate complex reasoning, including single-hop, multi-hop, and numerical reasoning. Additionally, InfographicVQA \cite{mathew2022infographicvqa} presents a diverse collection of infographics paired with question-answer annotations, establishing a rigorous benchmark to test multimodal document understanding. Finally, from a synthetic data perspective, the integration of synthetic data in training Deep Learning (DL) models for text \cite{jaderberg2014deep}, \cite{gurjar2018learning}, and handwritten text \cite{krishnan2016generating} recognition in natural images has reduced dependence on labor-intensive human labeling.
Additionally, it has boosted the capabilities of DL models, enabling scalability with an increase in the number of samples. Finally, Blender emerges as a pivotal tool for generating synthetic images to train DL models. Widely recognized for its versatility, it is extensively employed, showcasing its efficacy in creating synthetic image data for object detection \cite{mayershofer2021towards}, digital image correlation \cite{rohe2022generation}, or endoscopic datasets for validating surgical vision algorithms \cite{cartucho2020visionblender}. Furthermore, BlenderProc \cite{denninger2020blenderproc} has bridged the gap between synthetic training and real-world test domains in computer vision tasks. These findings collectively back Blender's use in synthetic sample generation.

From the Visually rich Document Understanding (VrDU) perspective, the LayoutLM family \cite{Xu2020} stands out. This model builds on top of BERT \cite{devlin2018bert}, but in addition to text, it also includes a multimodal input with layout and image (which Faster R-CNN \cite{ren2016faster} converts into visual embeddings). The first version of this family (LayoutLM) is pre-trained on tasks such as document classification and form understanding (as a key-value extraction task). On the other hand, LayoutLMv2 \cite{Xu2020a} introduces new pre-training tasks (text-image alignment and text-image matching) aimed at better capturing the image-text-layout interaction. Moreover, LayoutXLM \cite{Xu2021} is built on top of this model, striving to overcome language barriers by using a corpus with samples from 53 languages and providing the XFUND dataset \cite{Xu2022} as a benchmark. Afterward, LayoutLMv3 \cite{Huang2022} appears as a new family release. This model is the first one that does not use a CNN or RCNN to obtain the embeddings of the visual part. In addition, to achieve a better cross-modal representation, they include a Word-Patch Alignment task, intending to induce a correlation between an image fragment and its corresponding text fragment (here, LayoutLMv3 can only discriminate whether a patch is masked or not, not reconstruct it). Despite the promising results obtained by this family of models, there are friction points, such as their dependence on OCRs. This dependence translates into OCR difficulties when dealing with challenging real-world scenarios \cite{palacios2008system}, but also presents a more subtle challenge: the reading order of OCRs (which determines the input sequence of tokens to the model). XYLayoutLM \cite{gu2022xylayoutlm} highlights this dependence and proposes a token order correction based on the location of words (x, y coordinates). In line with the efforts to minimize OCR-related errors and computational costs in document understanding, Donut \cite{kim2021donut} represents a paradigm shift toward OCR-free models. Based on this end-to-end pipeline, DocParser \cite{dhouib2023docparser} improves results to achieve state-of-the-art results by better capturing discriminative character features. Finally, in terms of performance, Universal Document Pro-cessing (UDOP) \cite{tang2023unifying} is state of the art in up to 8 VrDU-based tasks. For the first time, a model includes editing and generating realistic documents during pre-training (going further than LayoutLMv3). In addition, this model also unifies the architecture into a single vision-text-layout transformer.

\section{Dataset generation and pipeline overview}
\label{sec:datasetGeneration}

\begin{figure}
\centering
\includegraphics[width=0.85\columnwidth]{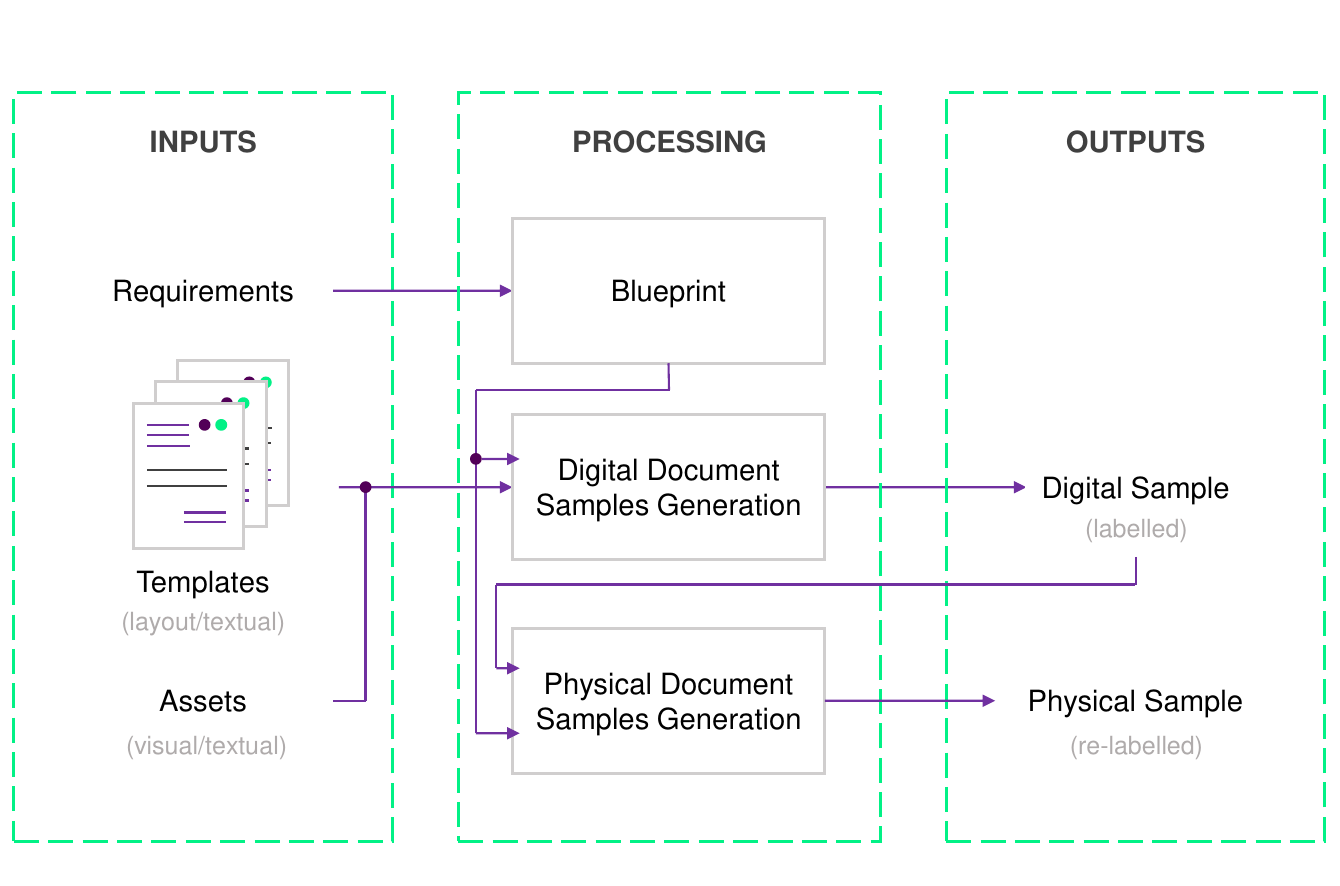}
\caption{Pipeline overview.} 
\label{fig:pipelineOverview}
\end{figure}

The MERIT Dataset's sample generation pipeline produces labeled images from school records. It can generate samples in different languages and schools depending on the user's necessities. The pipeline facilitates the generation of image samples in two distinct styles: digitally originated documents and documents set in photorealistic contexts, as Figure \ref{fig:digitalAndPhysicalSamples} shows. Figure \ref{fig:pipelineOverview} depicts an overview of the synthetic generation pipeline and its main components.

\subsection{Inputs}
\label{subsec:inputs}
The system requires users to provide a set of assets and configuration files as inputs, which are essential for the seamless operation of the automated pipeline.
    
\subsubsection{Requirements}
A configuration file that users fill out to detail functional aspects of their dataset, including a selection of schools, the number of students per school, and subjects per page in each template. This file also enables users to embed biases within their samples, such as gender ratios or the cultural origins of names. Besides, it allows the inclusion of grade biases based on these parameters to study biases in LLM models.

\subsubsection{Templates}
\label{subsec:pipepileOverviewTemplates}
\begin{figure}
\centering
\includegraphics[width=1\columnwidth]{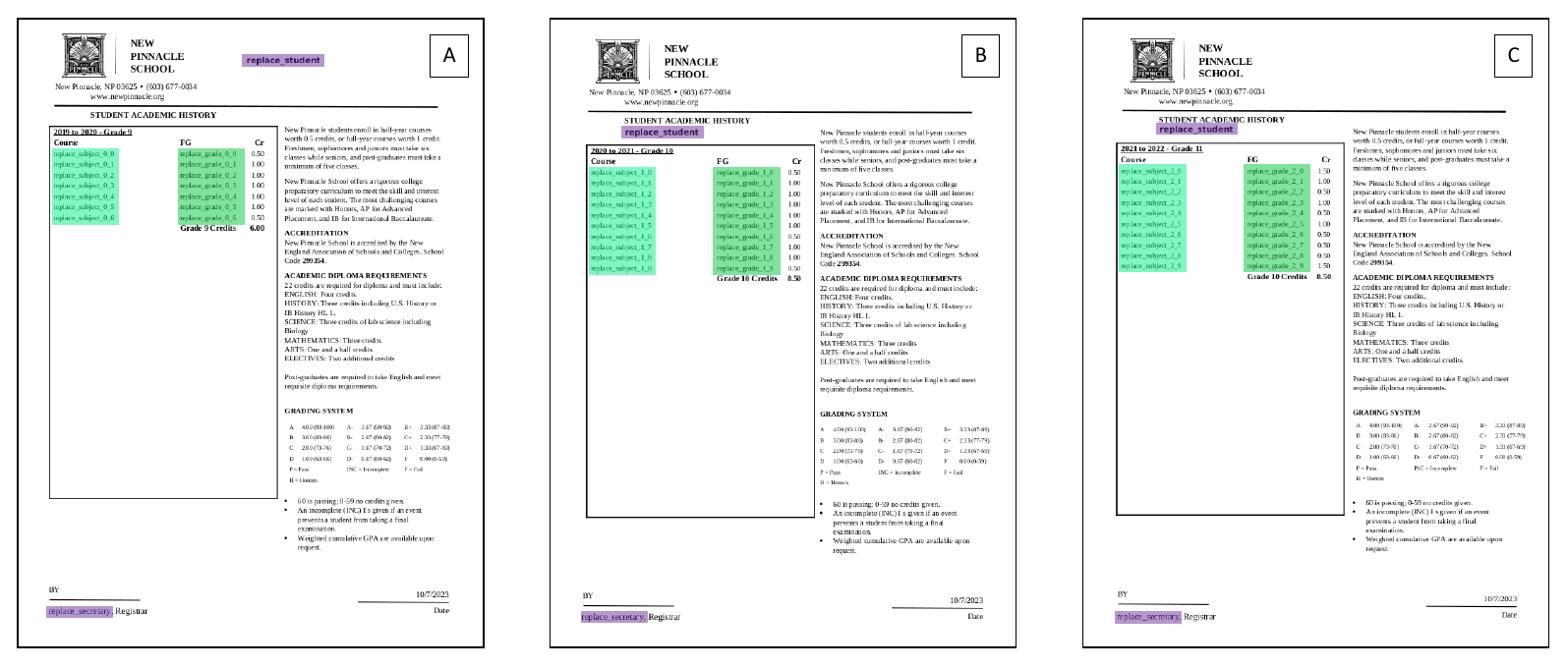}
\caption{A template example with highlighted keywords. Subjects and grades are highlighted in green, and name-related keywords are in purple.} 
\label{fig:templateSample}
\end{figure}

This research offers a dataset and a synthetic dataset generation pipeline, which includes an intuitive interface for template management. These templates, crucial for generating samples, dictate the layout and serve as the foundation for the samples' textual and visual elements. They contain replaceable keywords for dynamic content creation, such as the principal's name, secretary's name, student name, subject name, and corresponding grades, as shown in Subsection \ref{subsec:documentGeneration}. An example template, with replaceable keywords highlighted, is presented in Figure \ref{fig:templateSample}.

\subsubsection{Assets}
\begin{figure}
\centering
\includegraphics[width=1\columnwidth]{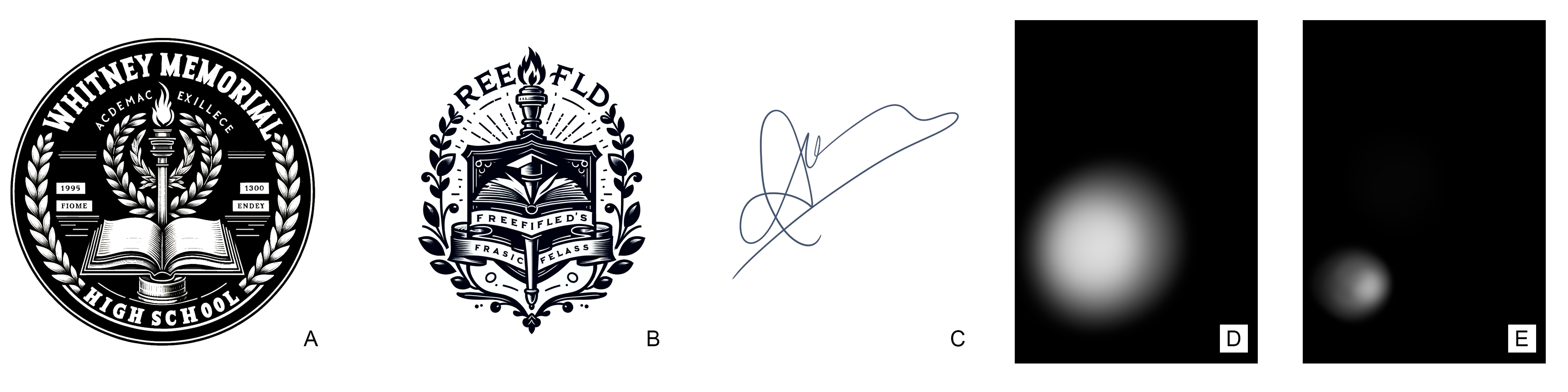}
\caption{Visual Assets: Every school template includes visual assets that help build its unique identity. The samples explicitly display some assets, such as stamps, badges, and signatures (A-C). Others assist in randomizing the positions of explicit assets, including maps for stamps and signatures (D and E).}
\label{fig:visualAssets}
\end{figure}

Assets enrich the sample generation process and are divided into textual and visual categories. Textual assets, comprising databases of names from 17 languages or regions and synonyms for subject names in 5 languages across 26 themes, allow for diverse and biased sample creation. The MERIT Dataset includes explicitly Spanish and English templates with names from 7 origins (see Section \ref{sec:datasetAnalysis} for further details). Visual assets, as illustrated in Figure \ref{fig:visualAssets}, include assets that either directly appear in the samples or assets designed to help position other assets (such as maps, which are probabilistic distributions defined as grayscale images). Visual assets include school stamps (A), badges (B), signatures (C), stamp maps (D), and signature maps (E). These assets are prepared for seamless integration and help create authentic-looking digital document samples.

\subsection{Blueprint}
\label{subsec:blueprint}
The blueprint serves as the central component of the pipeline, encompassing comprehensive details of all samples. It consists of various fields: informational (e.g., sample names), management-related (indicating the need for reprocessing in the photorealistic pipeline), and directly impactful ones that enhance flexibility and control over the dataset's content, like the origin of names, gender, or parameters for the student's grade note. Furthermore, the blueprint's creation is guided by the requirements file, a key element allowing user fine-tuning. This file dictates aspects such as sample language, involved schools, or student distribution based on gender and ethnic name origin and the possibility of including biases in grading (which implies that the MERIT Dataset and its generation pipeline is a great tool to benchmark LLM ethics and potential biases).

\subsection{Digital document samples}
\label{subsec:documentGeneration}

\begin{figure}
\centering
\includegraphics[width=0.85\columnwidth]{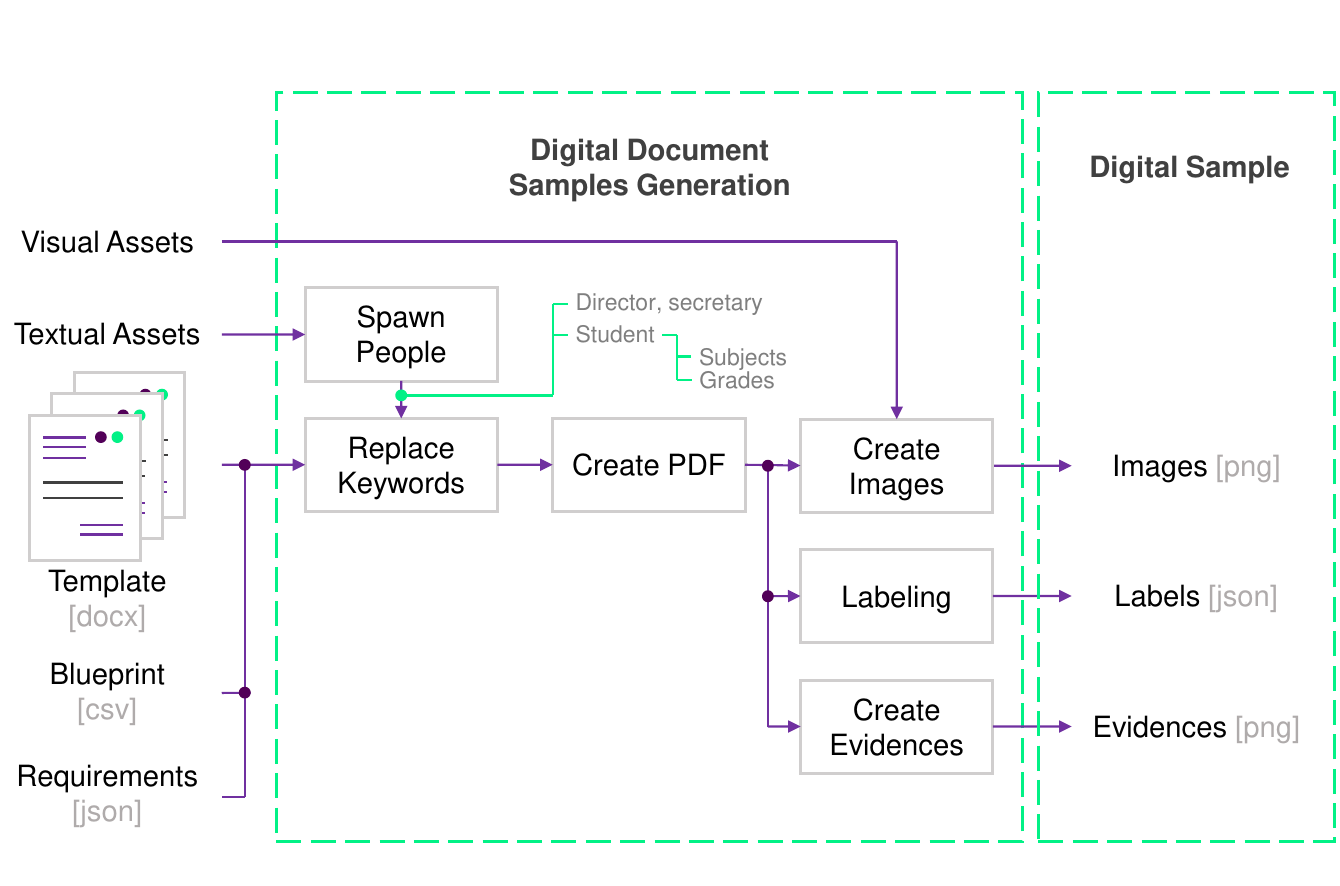}
\caption{Pipeline detail for generating Digital Document Samples.} 
\label{fig:pipelineDigitalSamples}
\end{figure}

The Digital Sample generation module is the initial phase of the pipeline. It is crucial in creating digital document images and their corresponding text and labels. Figure \ref{fig:pipelineDigitalSamples} provides a detailed overview of the components within this module. At the heart of this module is creating people instances, which are then used to populate the templates. After generating these profiles, the module leverages methods to replace keywords or produce evidence to ensure the labeling quality.

\subsubsection{People spawning}
\label{subsubsec:spawnPeople}

As detailed in Section \ref{subsec:pipepileOverviewTemplates}, templates are designed with keywords that the pipeline must dynamically replace. Some of these keywords denote individuals and bifurcate into two categories: administrative personnel (principals or secretaries) and students.

The process is straightforward for administrative roles involving only a name attribute. Name generation is based on a random selection method that selects names from the relevant language database, ensuring that administrative identities remain consistent across student samples within the same school.

Student instances introduce a higher level of complexity, incorporating the student's name and academic record (subjects and grades) as an additional attribute.

\begin{itemize}
    \item Names. Student names are generated based on user-defined parameters in the requirements file. Users can specify both the gender (male or female) and the origin of the name, with the pipeline offering the flexibility to sample from as many as 17 different languages or origins.

    Furthermore, users can also set the probability of generating male or female students and select the likelihood that a name of a particular origin is generated. Following these determinations, student names are generated similarly to the process described for administrative roles.

    \item Subjects. The curriculum for each course is predefined by the template model, which sets the number of subjects. For each student's course, subjects are selected through a randomized process from the subject database corresponding to the template language. This process involves two steps: initially, a subject topic is chosen from among 26 available themes per language (e.g., \textit{mathematics}). Once the subject topic is selected, a specific subject synonym is selected by sampling the available options (such as \textit{calculus}, \textit{trigonometry}, or \textit{algebra} for the \textit{mathematics} theme). This selection of synonyms is also randomized, following a uniform distribution.

    \item Grades. The process of grade generation serves as the focal point for potential bias introduction within the dataset. Users have the discretion to define the parameters (namely, mean and standard deviation) that shape the normal distributions, thereby modulating student grades by gender and name origin. As the creators of this dataset, we consider this capability crucial for identifying biases, implementing corrective measures, or simply acknowledging the existence of biases when deploying LLMs. For instance, we can prompt an LLM to select a subset of candidates from a highly heterogeneous pool of students of different gender and backgrounds. Given that our dataset is fully labeled and traceable, we can monitor whether the LLM relies on objective data (grades) to make candidate selections or biases learned during pre-training (i.e., the LLM selects candidates based on other factors).
    
\end{itemize}

\subsubsection{Keyword replacement and PDF creation}
\label{subsubsec:replaceAndCreatePDF}
The Replace Keywords module substitutes predefined keywords within templates, utilizing an XML file as the source from which the DOCX format template is derived. Following the generation of the DOCX file with the information now replaced, the conversion to a PDF format is seamlessly executed.

\subsubsection{Generation of image samples}
\label{subsubsec:createImages}
Image generation in standard formats like PNG is efficiently achieved using libraries such as PIL. As illustrated in Figure \ref{fig:pipelineDigitalSamples}, this process is enhanced by incorporating visual assets like signatures and stamps to bolster the documents' realism. Realism in human actions often stems from imperfection, so signatures and stamps include corresponding heatmaps. These grayscale maps, sized to match the generated samples, depict the probability of placing the associated asset in a particular location—the lighter the pixel, the higher the probability. This approach for determining asset placement, along with slight randomized rotations and scaling adjustments for signatures, accurately mimics the human act of stamping and signing documents. Moreover, this technique prevents visual models from relying on a static reference, which could lead to the formation of unreliable patterns based on graphical references.

\subsubsection{Labeling}
\label{subsubsec:labeling}
The annotating process is one of the most significant contributions of this work. Leveraging pre-configured assets in text and layout, our pipeline can produce labeled samples that precisely align the visual, textual, and layout elements. This process is fully automated once the assets are correctly configured, removing any marginal generation cost in human time.

Regarding the labeling format, our pipeline adheres to the FUNSD Dataset \cite{Jaume2019} format, which is directly applicable, for instance, in models like LayoutLM. Our dataset achieves a level of labeling detail and granularity beyond previous datasets. For example, whereas the FUNSD dataset (a widely used benchmark for testing the capabilities of the LayoutLM model family) offers labels like `other,' `question,' `answer,' and `heading,' our dataset encompasses an array of 26 subject themes, presented in two languages, for four distinct educational levels (serving as the `question' label), and their corresponding grades (serving as the `answer' label), along with the `other' label for text deemed non-relevant. This brings the total to 417 distinct labels. Thus, we assert that the MERIT Dataset elevates the complexity of tasks such as VrDU or Key Information Retrieval, challenging models to discern much subtler characteristics of layout, text, and visual cues to accomplish the task.

The structure and content of the labels are organized into segments (groups of words limited to the length of a line). Each segment includes an associated label and a bounding box defining its location using two points: the top-left and bottom-right corners, under the assumption of orthogonality. Moreover, the labels nest all the words constituting the segment and their bounding boxes.

\subsubsection{Evidence creation}
\label{subsec:createEvidence}
\begin{figure}
\centering
\includegraphics[width=1\columnwidth]{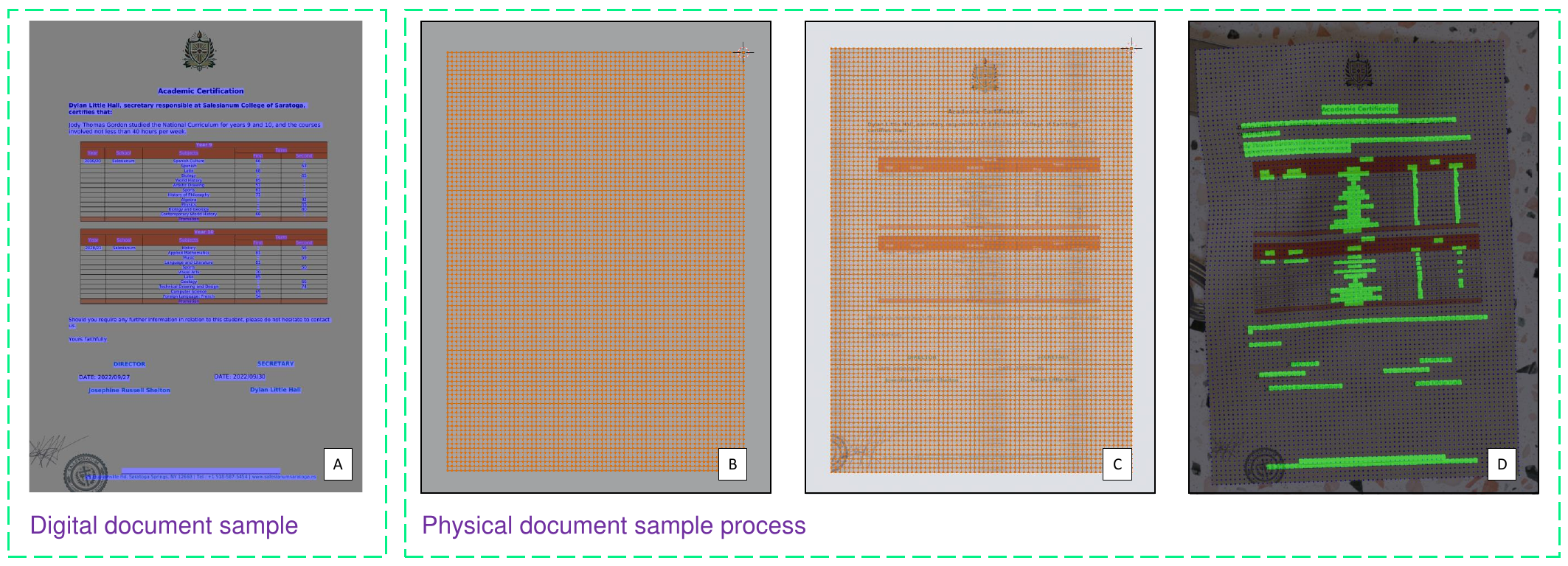}
\caption{Bounding boxes tracking. Visual evidence of a digital document sample (before Blender) with highlighted words bounding boxes (A), quadrilateral regular mesh in Blender (B), overlay of mesh and texture (C), and visual evidence after cloth simulation (D).}
\label{fig:verticesAndMesh}
\end{figure}

This module generates visual evidence to facilitate error debugging, verify the labeling process's precision, and ensure that the labels' bounding boxes match the correct regions within the image. It also manages internal parameters to scale PNG dimensions when creating them from PDF files. Figure \ref{fig:verticesAndMesh}.A showcases an example of this evidence, displaying a sample with highlighted bounding boxes.

\subsection{Physical document sample: photorealism in Blender}
\label{subsec:photorrealismBlender}

\begin{figure}
\centering
\includegraphics[width=1\columnwidth]{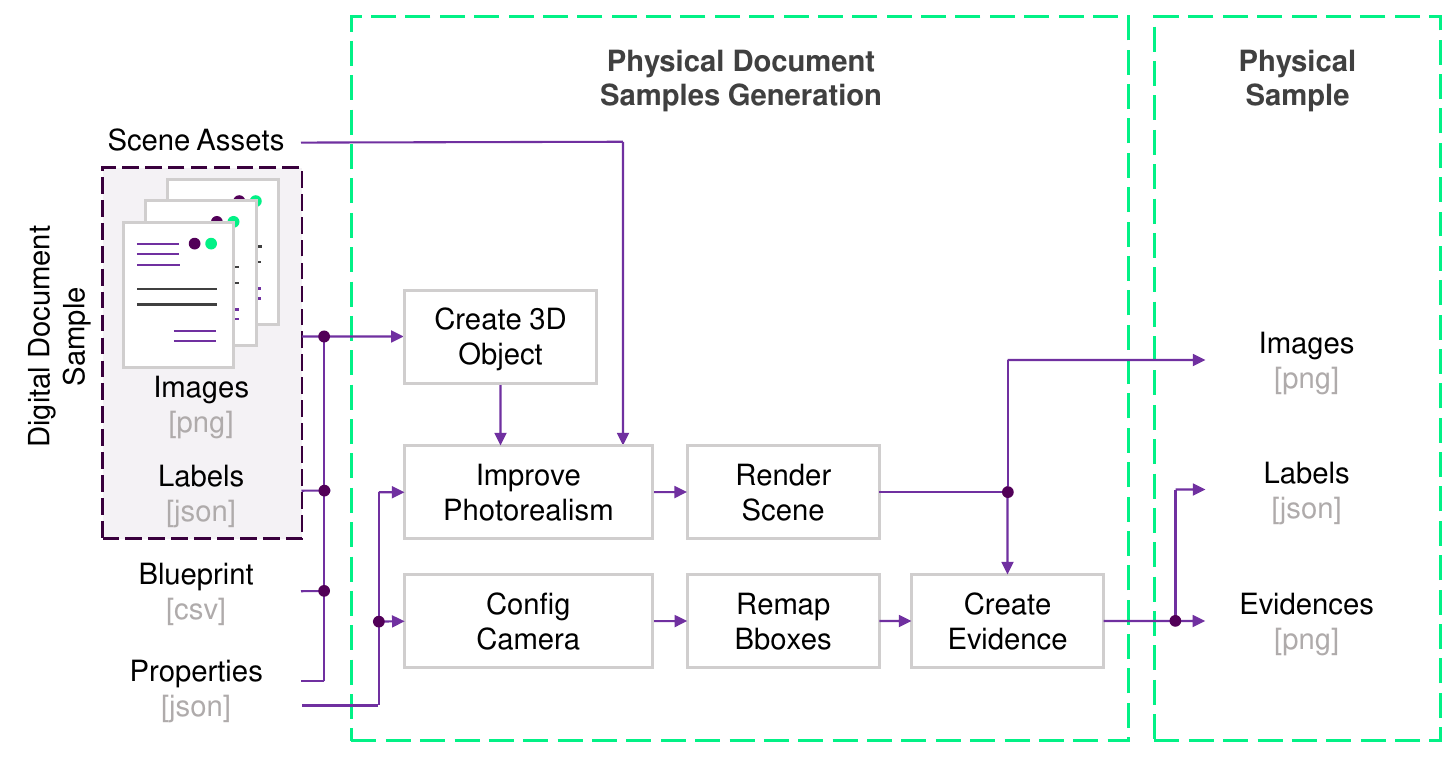}
\caption{Pipeline detail for generating Physical Document Samples.}
\label{fig:pipelinePhysicalSamples}
\end{figure}

The Physical Document Sample Generation module specializes in the visual transformation of samples created by the Digital Document Samples module discussed in Section \ref{subsec:documentGeneration}. This module does not modify the layout or textual content but focuses on augmenting the original documents' visual attributes. This pipeline section automatically provides the images with photorealistic features, including lighting, background, and camera settings management. Figure \ref{fig:pipelinePhysicalSamples} shows the components of this module, further elaborated in the following subsections.

\subsubsection{3D object creation}
\label{subsubsec:3dObjectCreation}
The main object to model in the 3D scene is a sheet of paper, defined as a plane with the proportions of a DIN A4 sheet. Initially, this plane is defined by four vertices. However, defining key points as part of the paper's mesh helps facilitate the tracking of word-bounding boxes when employing a camera in the scene. In addition, more detailed meshes are beneficial for applying cloth simulation.

The process begins by defining a smooth quadrilateral mesh. Once this is done, the corners of words' bounding boxes from the Digital Samples generator (Section \ref{subsec:documentGeneration}) are approximated to the paper's mesh vertices so that every bounding box point is mapped to a vertex in the quadrilateral mesh. Although a Delaunay triangulation \cite{lee1980two} might be a more elegant solution for adapting the bounding box vertices to the original plane, we have empirically proved its incompatibility with smooth cloth simulation results in Blender.

Once the mesh is ready, we overlay the original document image (output image shown in Figure \ref{fig:pipelineDigitalSamples}) onto its top face. With an appropriate mesh and texture, the pipeline is set to apply additional modifications to enhance the scene's photorealism (lighting, imperfections, etc.). Figure \ref{fig:verticesAndMesh} displays the mesh generated from the vertices (B), a mesh overlay on the texture to confirm its fit within the bounding boxes (C), and visual evidence demonstrating proper bounding box tracking in Blender after cloth simulation (D).

\subsubsection{Photorealism improvements}
\label{subsubsec:photorealismImprovemnents}
Achieving photorealism in static synthetic images is possible with accurate mesh modeling, realistic texture design, and adding imperfections that naturally occur through human interaction and environmental factors. We identified key and recurring factors that characterize images in this context by studying real samples from university admission processes. Based on these observations, the dataset incorporates the following conditions:

\begin{itemize}
    \item Lightning Conditions. The scene's lighting conditions significantly vary the image, creating difficulties for textual information extraction due to low light or overexposure. It's common to find documents scanned or photographed under artificial lighting or in natural, diffused light conditions.
    
    \item Background. Real-world images often feature desks as the background, supporting the photographed paper. Moreover, it's typical for these images to extend beyond the primary area of interest (the paper), capturing additional objects like office supplies. Incorporating variations in the background is empirically beneficial and serves as a common strategy in model training with synthetic images. This approach helps to narrow the gap between the distributions of real and synthetic images, enhancing model performance when training with synthetic datasets and inferring with real images.
    
    \item Paper textures. Paper textures introduce physical world imperfections, such as the fibers of organic material like paper, folds and wrinkles, and even stains from human handling. This pipeline features 13 different paper textures and methods for generating stains typical of printing and scanning documents.
    
    \item Shadow Casting. Shadows appear on photographed objects when a person blocks the light source, for instance, when taking a picture with a mobile device. The pipeline incorporates an articulated human model to simulate shadows in Blender. The model's position is randomized to position it between the light source and the document consistently.

\end{itemize}

\subsubsection{Camera configuration and scene renderization}
\label{subsubsec:cameraConfigAndRender}
Creating photorealistic images from a scene is significantly influenced by the rendering engine settings (EEVEE for this pipeline) and the camera's settings. Tables \ref{table:eevee_render_settings} and \ref{table:camera_settings} provide configuration insights for the rendering engine and camera, respectively.

\begin{table}
\centering
\begin{minipage}[b]{0.45\textwidth}
\centering
\small
\textbf{Rendering Engine Config}\\[2ex]
\begin{tabular}{ll}
\hline
\textbf{Setting}                   & \textbf{Value}    \\ \hline
\multicolumn{2}{c}{Shadows}                  \\ \hline
Soft Shadows                       & True              \\
Shadow Threshold                   & 0.01              \\ \hline
\multicolumn{2}{c}{Indirect Lighting}         \\ \hline
Reflection Cubemap Size            &                   \\
Irradiance Volume                  &                   \\
Bake Indirect Lighting             &                   \\ \hline
\multicolumn{2}{c}{Samples and Denoising}     \\ \hline
Samples                            & 64                \\
Denoising                          & True              \\ \hline
\end{tabular}
\caption{EEVEE Rendering Settings}
\label{table:eevee_render_settings}
\end{minipage}
\hfill
\begin{minipage}[b]{0.4\textwidth}
\centering
\small
\textbf{Camera Config}\\[2ex]
\begin{tabular}{ll}
\hline
\textbf{Setting}                   & \textbf{Value}                               \\ \hline
\multicolumn{2}{c}{Camera}                                          \\ \hline
Location X                         & 0.105 m                         \\
Location Y                         & 0.1485 m                        \\
Location Z                         & $\mathcal{N}$(0.55, 0.05) m                       \\ \hline
Rotation X                         & $\mathcal{N}$(0, 1) º                            \\
Rotation Y                         & $\mathcal{N}$(0, 4) º                            \\
Rotation Z                         & $\mathcal{N}$(180, 5) º                          \\ \hline
\multicolumn{2}{c}{Depth of Field}                                     \\ \hline
F-Stop                             & 2.8                                           \\ \hline
\multicolumn{2}{c}{Lens}                                                   \\ \hline
Focal Length                       & 50 mm                                         \\
Lens Type                          & Perspective                                   \\
Sensor Size                        & 36 mm                                         \\ \hline
\end{tabular}
\caption{Camera Settings}
\label{table:camera_settings}
\end{minipage}
\end{table}

\subsubsection{Remapping bounding boxes and evidence creation}
\label{subsubsec:remapBboxesAndEvidenceCreation}
Table \ref{table:camera_settings} illustrates that the camera's position and orientation vary, randomized according to normal distributions. These 3D parameters, along with the focal length or sensor size, force the original document's bounding boxes to be mapped and transformed to the newly rendered image's coordinates. Blender's predefined functions ease the mapping process. By defining a mesh with vertices positioned at the original bounding boxes' locations (as described in Section \ref{subsubsec:3dObjectCreation}), it is possible to track these vertices to the new coordinates in the image. Following the retrieval of these coordinates, just as exposed in Section \ref{subsec:createEvidence}, images are generated to serve as evidence, aiding in debugging the layout labeling process. Figure \ref{fig:verticesAndMesh}.D shows an evidence image after the Blender transformation pipeline. Finally, the label file is updated to reflect the new bounding box values.

\section{Dataset analysis}
\label{sec:datasetAnalysis}
The MERIT Dataset is a synthetic dataset of labeled images created to push the limits of Visually-rich Document Understanding. It was generated using the pipeline described in Section \ref{sec:datasetGeneration}. The dataset consists of 33k samples, with each original sample corresponding to a processed sample in Blender. It includes documents in English and Spanish, using seven different school templates per language. In the following subsections, we explain the dataset's folders and data structure, and we point out its main features, which are divided into four categories: layout, text, visual, and ethical features.

\subsection{Dataset structure}
\begin{figure}
\centering
\includegraphics[width=1\columnwidth]{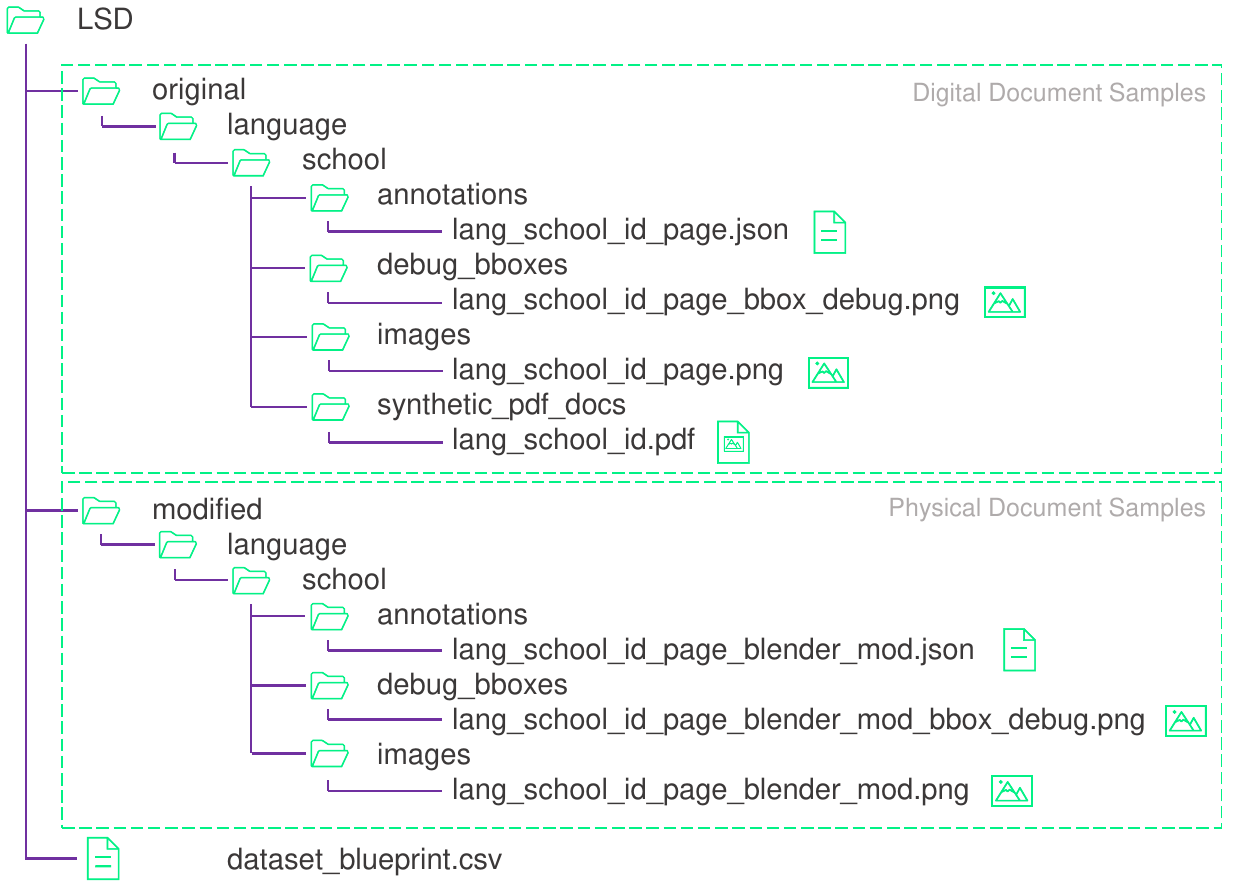}
\caption{Dataset Structure.}
\label{fig:datasetStructure}
\end{figure}

Figure \ref{fig:datasetStructure} displays the folder structure of the MERIT Dataset. This structure divides the data into two main sections: data generated as Digital Samples (Section \ref{fig:pipelineDigitalSamples}) and data generated as Physical Samples (Section \ref{fig:pipelinePhysicalSamples}). This division enhances traceability and ensures access to the original images, even though a portion has been modified using the Blender block. The \textit{language} folders are specific to each language (Spanish and English). Similarly, each \textit{school} folder is dedicated to an individual school. In addition to the target images, labels, and debug images, the dataset also retains the original PDF documents.

Table \ref{table:datasetGeneralStatistics} presents the dataset's contents, listing the total number of samples categorized by school and language. Each school consists of 1k students, and the school's template determines the number of samples per student.

\begin{table}[ht]
\centering
\small

\begin{minipage}[t]{0.45\textwidth}
\centering
\begin{tabular}{llll}
\multicolumn{4}{c}{\textbf{English}} \\ \hline
Lang. & School & Tag & Samples \\ \hline
Eng. & Freefields & A & 3000 \\
 & Greenfields & B  & 3000  \\
 & James & C  & 1000  \\
 & Paloalto & D  & 2000  \\
 & Pinnacle & E  & 3000  \\
 & Salesianum & F  & 2000  \\
 & Whitney & G & 2000  \\
\hline
& & & 16000 \\
\end{tabular}
\end{minipage}
\hfill
\begin{minipage}[t]{0.45\textwidth}
\centering
\begin{tabular}{llll}
\multicolumn{4}{c}{\textbf{Spanish}} \\ \hline
Lang. & School & Tag & Samples \\ \hline
Spa. & Aletamar & H & 2000 \\
 & Británico & I  & 3000  \\
 & Deus & J  & 2000  \\
 & Liceo & K  & 3000  \\
 & Lusitano & L  & 2000  \\
 & Monterraso & M  & 2000  \\
 & Patria & N  & 3000  \\
\hline
& & & 17000 \\
\end{tabular}
\end{minipage}

\caption{MERIT Dataset general statistics. The tag column refers to tags used in Figure \ref{fig:datasetVisualOptionsDigitalSamples} to identify samples.}
\label{table:datasetGeneralStatistics}
\end{table}

\begin{figure}
\centering
\includegraphics[width=1\columnwidth]{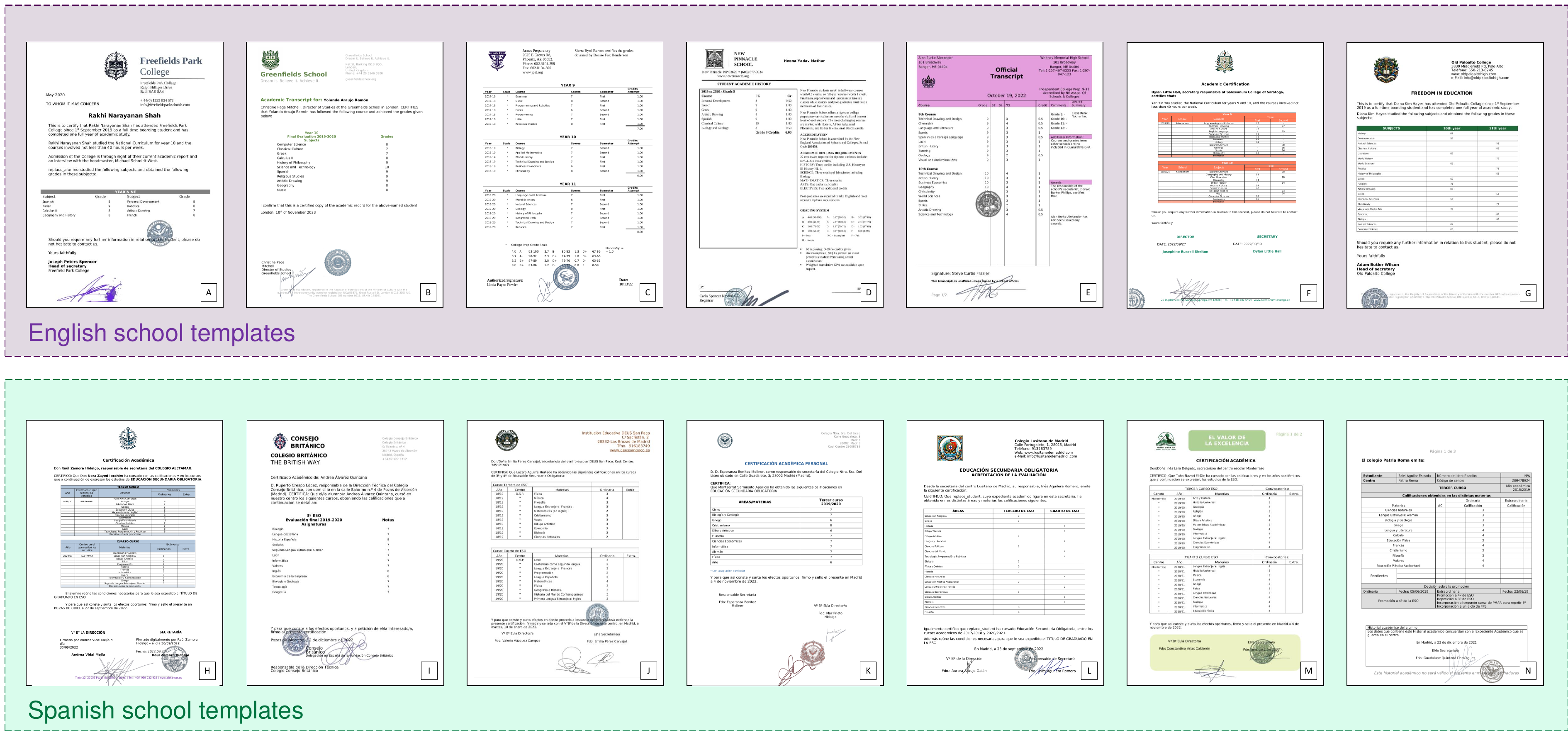}
\caption{Layout and visual aspects of the dataset digital samples, divided by school. Samples in English are A-E and samples in Spanish are F-L.}
\label{fig:datasetVisualOptionsDigitalSamples}
\end{figure}

\subsection{Layout features}
\begin{figure}
\centering
\includegraphics[width=1\columnwidth]{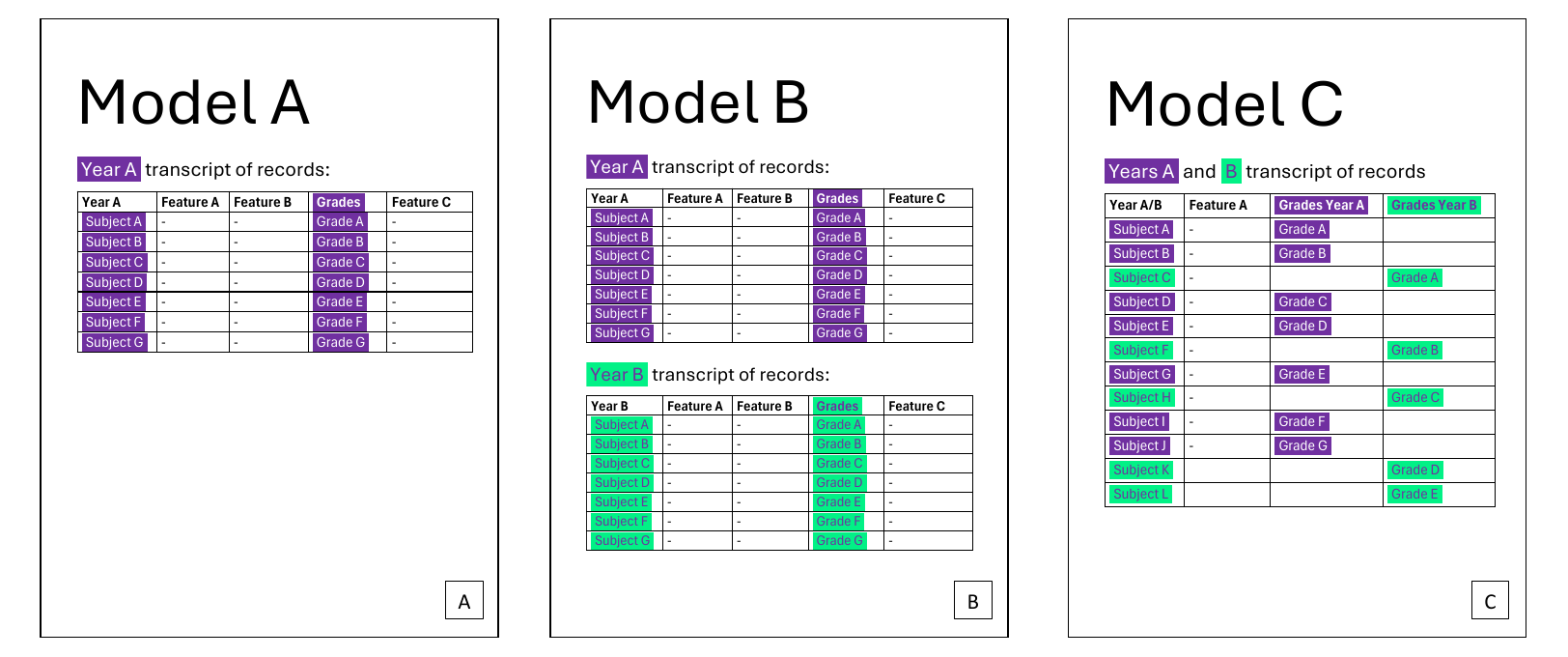}
\caption{Layout structures considered in the dataset.}
\label{fig:layoutFeatures}
\end{figure}

The MERIT Dataset features samples with distinct layout patterns, which remain consistent across all samples from a particular school, ensuring each student's sample is unique; no two students share the same document. The dataset comprises three primary layout models, each with slight variations. This variety ensures broad coverage of real-world scenarios found in school records. Moreover, this layout diversity elevates the challenge for models engaged in the VrDU task, effectively bridging the gap between synthetic and real sample distributions. Figure \ref{fig:layoutFeatures} illustrates the three main layout models:

\begin{itemize}
    \item Model A: This model features a standalone table for each grade level, one per page, making it the most straightforward layout since it avoids mixing information from different grades. It typically has one column for subjects and another for grades, with a more complex variant that features two columns for subjects and two for grades.
    
    \item Model B: Features individual tables for each grade level, with more than one table per page.

    \item Model C: Incorporates a single table that accommodates two grade levels per page, representing the most complex variation. It includes one column for subjects and two columns for grades, with each set of grades corresponding to a different grade level. This model demands precise attention to layout, as the models must accurately associate text positions with specific columns and rows to correctly label words of interest. The challenge is more significant in scenarios where the Blender module's camera position adjustments result in the non-orthogonal alignment of table columns and rows with the PNG margins.
\end{itemize}

Table \ref{table:layoutStatistics} lists the number of samples for each layout model and their respective proportions within the dataset.

\begin{table}[ht]
\centering
\small
\begin{tabular}{llll}
\hline
Layout & Variation & Samples & Fraction \\ \hline
Model A & Single column & 22k & 68.75\% \\
Model A & Double column & 3k & 9.38\%  \\
Model B & Two tables & 5k & 15.63\%  \\
Model B & Three tables & 1k & 3.13\%  \\
Model C & --- & 2k & 6.25\%  \\
\hline
\end{tabular}
\caption{Layout features statistics for the MERIT Dataset.}
\label{table:layoutStatistics}
\end{table}

\subsection{Visual features}

The dataset consists of 33k digital document samples (original samples). The visual content of these samples is shaped by the visual features introduced in the templates (along with the randomization of visual assets for position, orientation, and size where applicable). As the visual aspect of the digital samples is greatly determined by the template, the figures presented in Table \ref{table:datasetGeneralStatistics} are also valid statistics to describe the visual content of the digital samples. Accordingly, Figure \ref{fig:datasetVisualOptionsDigitalSamples} displays the visual appearance of representative samples from each school and reiterates the figures from Table \ref{table:datasetGeneralStatistics} to enhance readability.

\begin{figure}
\centering
\includegraphics[width=1\columnwidth]{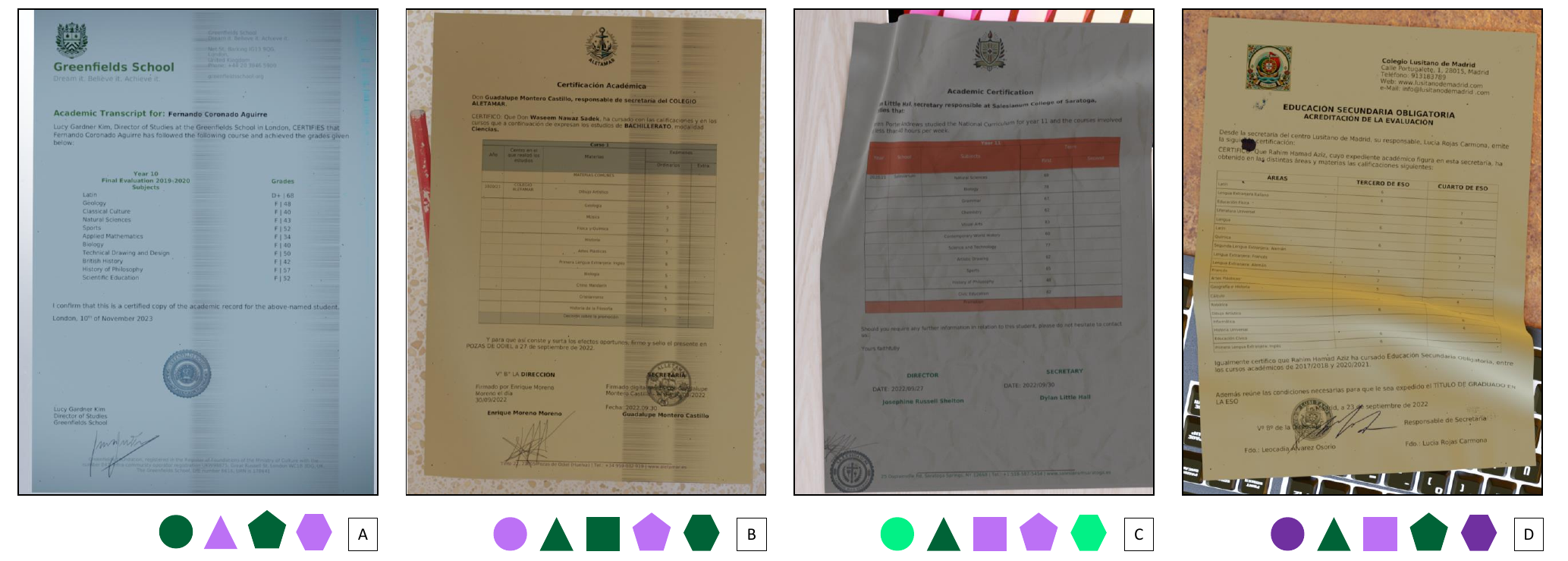}
\caption{Visual styles achieved after Blender module transformations.}
\label{fig:datasetBlenderModStyles}
\end{figure}

All 33k original samples have been enhanced through Blender's photorealistic module. Detailed in Section \ref{subsubsec:photorealismImprovemnents}, this module performs several operations that endow digital samples with new visual features, steering them towards a more photorealistic appearance. Table \ref{table:datasetVisualStatistics} outlines the distribution of these enhanced samples (Physical Document Samples) according to the applied modifications. Furthermore, Figure \ref{fig:datasetBlenderModStyles} illustrates examples of the distinct visual styles achieved by the modifications detailed in Table \ref{table:datasetVisualStatistics}.

\begin{table}[ht]
\centering
\small
\begin{tabular}{lllll}
\hline
Feature & Option & Tag & Samples & Fraction \\ \hline
Rendering style & Scanner & {\tikz\fill[premierEmerald+] (0,0) circle (0.75ex);} & 10105 & 30.62\% \\
& Natural & {\tikz\fill[premierPurple-] (0,0) circle (0.75ex);} & 8283 & 25.10\%  \\
& Studio & {\tikz\fill[premierEmerald] (0,0) circle (0.75ex);} & 8100 & 24.55\%  \\
& Warm & {\tikz\fill[premierPurple] (0,0) circle (0.75ex);} & 6512 & 19.73\% \\ 
\hline
Mesh modification & Cloth simulation & {\begin{tikzpicture}\fill[premierEmerald+] (0,0) -- (1.5ex,0) -- (0.75ex,1.5ex) -- cycle;\end{tikzpicture}} & 20371 & 61.73\% \\
& Simple plane & {\begin{tikzpicture}\fill[premierPurple-] (0,0) -- (1.5ex,0) -- (0.75ex,1.5ex) -- cycle;\end{tikzpicture}} & 12629 & 38.27\% \\ \hline
Shadow casting & True & {\tikz\fill[premierEmerald+] (0,0) rectangle (1.5ex,1.5ex);} & 21688 & 65.72\% \\ 
& False & {\tikz\fill[premierPurple-] (0,0) rectangle (1.5ex,1.5ex);} & 11312 & 34.28\% \\ \hline
Background noise& Empty background & {\begin{tikzpicture}[baseline=-0.75ex]\fill[premierEmerald+] (-54:1ex) -- (18:1ex) -- (90:1ex) -- (162:1ex) -- (234:1ex) -- cycle;\end{tikzpicture}}& 20371 & 61.73\% \\ 
& Object in background & {\begin{tikzpicture}[baseline=-0.75ex]\fill[premierPurple-] (-54:1ex) -- (18:1ex) -- (90:1ex) -- (162:1ex) -- (234:1ex) -- cycle;\end{tikzpicture}} & 12629 & 38.27\% \\ \hline
Background material & Tiles & {\begin{tikzpicture}[baseline=-0.75ex]\fill[premierEmerald+] (0:1ex) -- (60:1ex) -- (120:1ex) -- (180:1ex) -- (240:1ex) -- (300:1ex) -- cycle;\end{tikzpicture}} & 13016 & 39.44\% \\ 
& Plastic & {\begin{tikzpicture}[baseline=-0.75ex]\fill[premierPurple-] (0:1ex) -- (60:1ex) -- (120:1ex) -- (180:1ex) -- (240:1ex) -- (300:1ex) -- cycle;\end{tikzpicture}} & 10105 & 30.62\% \\
& Wood & {\begin{tikzpicture}[baseline=-0.75ex]\fill[premierEmerald] (0:1ex) -- (60:1ex) -- (120:1ex) -- (180:1ex) -- (240:1ex) -- (300:1ex) -- cycle;\end{tikzpicture}} & 6578 & 19.93\% \\
& Metal & {\begin{tikzpicture}[baseline=-0.75ex]\fill[premierPurple] (0:1ex) -- (60:1ex) -- (120:1ex) -- (180:1ex) -- (240:1ex) -- (300:1ex) -- cycle;\end{tikzpicture}} & 3301 & 10.00\% \\
\end{tabular}
\caption{Visual features statistics for the MERIT Dataset. The tag column refers to tags used in Figure \ref{fig:datasetBlenderModStyles} to identify visual features in the subfigures.}
\label{table:datasetVisualStatistics}
\end{table}

\subsection{Textual features}
The information in Table \ref{table:datasetGeneralStatistics} is relevant for the dataset's general metrics and its textual dimensions: the dataset comprises 16k English and 17k Spanish samples. The textual content within the MERIT Dataset samples is derived from replicating and anonymizing actual student records, ensuring high realism in the textual content.

As detailed in Section \ref{subsubsec:spawnPeople}, the strategic management and automatic replacement of keywords introduce textual variability into the dataset. This approach ensures that the dataset encompasses a broad spectrum of textual information about students, as listed in Table \ref{table:keywordsTextualFeatures}.

\begin{table}[ht]
\centering
\small
\begin{tabular}{llllll}
\hline
Language & Gender & Origin & Students & Samples & Fraction \\ \hline
English & Female & English & 2476 & 5636 & 17.08\% \\
&& Spanish & 345 & 791 & 2.40\% \\
&& Chinese & 356 & 793 & 2.40\% \\
&& Indian & 354 & 825 & 2.50\% \\
\cline{2-6}
& & & 3531 & 8045 & 24.38\% \\
\cline{2-6}
& Male & English & 2436 & 5573 & 16.89\%\\
& & Spanish & 384 & 871 & 2.64\% \\
& & Chinese & 329 & 756 & 2.29\% \\
& & Indian & 320 & 755 & 2.29\% \\
\cline{2-6}
& & & 3469 & 7955 & 24.11\% \\
\hline
& & & 7000 & 16000 & 48.48\% \\ \hline
Spanish & Female & Spanish & 2423 & 5881 & 17.82\% \\
&& Arabic & 719 & 1755 & 5.32\% \\
&& Subsaharian & 365 & 885 & 2.57\% \\
\cline{2-6}
& & & 3507 & 8521 & 26.63\% \\
\cline{2-6}
& Male & Spanish & 2426 & 5891 & 18.41\% \\
&& Arabic & 716 & 1740 & 5.44\% \\
&& Subsaharian & 351 & 848 & 2.65\% \\
\cline{2-6}
& & & 3493 & 8479 & 25.69\% \\
\hline
& & & 7000 & 17000 & 51.51\% \\
\end{tabular}
\caption{MERIT Dataset: statistics for students and samples. Fractions refer to the total number of samples, not students.}
\label{table:keywordsTextualFeatures}
\end{table}

In addition to the textual content generated through keyword replacement for students, subjects, and grades, the dataset encompasses various textual attributes embedded within the school templates. These attributes contribute to the dataset's richness and diversity, including features such as grades represented as numbers or letters, dates, and page numbers.

\subsection{Ethical features: biases}
Given the ability to associate a student's name with their grade, bias could be introduced into the dataset. Despite the pipeline's intention to incorporate specific biases to shed light on the behaviors of commonly used models like ChatGPT, we have opted to release the MERIT Dataset, which relies on the most objective data for generating grades. Details on how these grades were determined for the templates in English and Spanish are outlined in \ref{annex:gradeBiases}. Parameters related to the origin of the name and the gender of the students are documented in Table \ref{table:origin_sex_grade_seeds}.

\begin{table}[htbp]
\centering
\small

\begin{tabular}{llll}
\multicolumn{4}{c}{\textbf{Name Origin}}         \\ \hline
Lang. & Origin & Country & $\mu$ \tablefootnote{Grades in Table \ref{table:origin_sex_grade_seeds} are scaled to addapt to different academic systems. For example, grades are scaled from 0-100 in the USA scenario. In some MERIT school samples these grades may be further remapped to a letter scale, ranging from F to A} \\ \hline
Eng. & English & USA & 68\\
& Spanish & El Salvador & 46 \\
& Chinese & China & 76  \\
& Indian & India & 44 \\ \hline
Spa. & Spanish & Spain & 6.54  \\
& Arabic & Morocco & 4.56  \\
& Subsaharian & Senegal & 4.06 \\ \hline
\multicolumn{4}{c}{\textbf{Name Gender}}         \\ \hline
Gender & & & $\mu$ \tiny{(USA/Spain)} \\ 
\hline
Female & & & 66/6.60\\
Male & & & 65/6.50\\
\hline
\end{tabular}
\caption{Parameters for sampling the normal distributions to generate student's grades in English and Spanish due to bias triggers: name origin and gender of the student. Student grades are determined by combining samples from two normal distributions: one linked to the student's name origin and the other to their gender. These parameters represent the mean values of the distributions on a scale of 0-10, with standard deviations set to 2 points in each case.}
\label{table:origin_sex_grade_seeds}
\end{table}

\begin{figure}
\centering
\includegraphics[width=1\columnwidth]{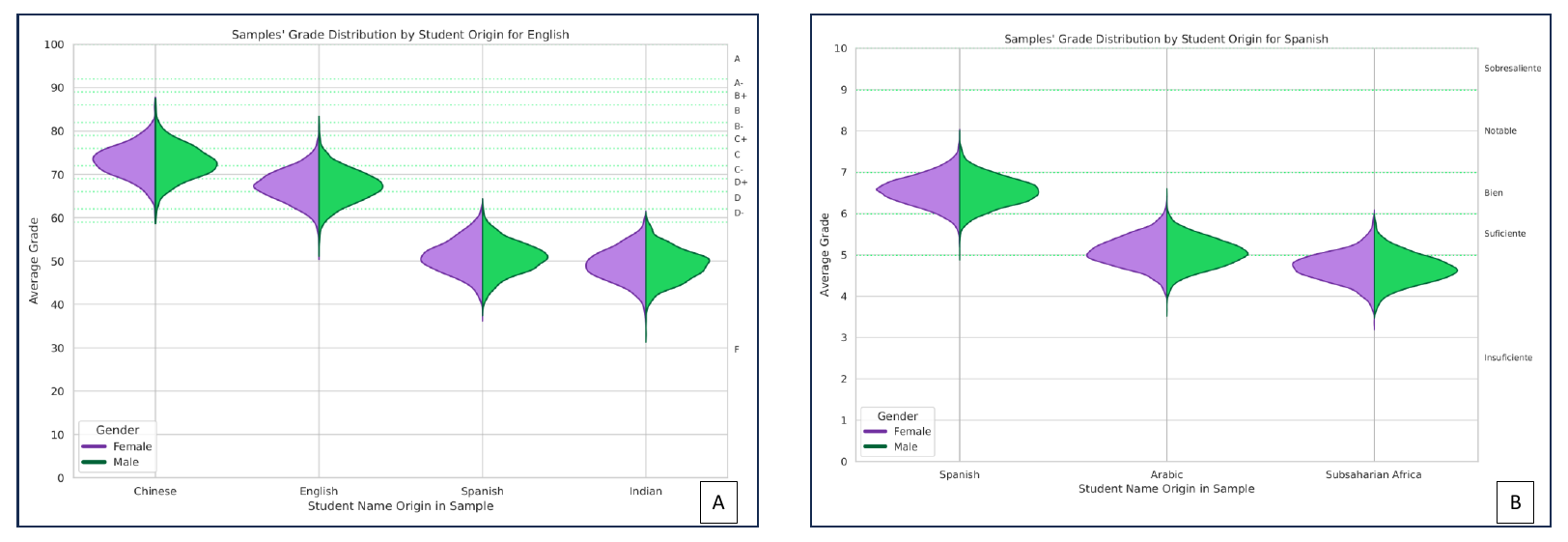}
\caption{Grade Distribution in the Dataset Based on PISA Reports (\ref{annex:gradeBiases}). Grades are categorized into two educational systems: the USA system (A) and the Spanish system (B). Within each system, grades are segmented by the gender and origin associated with the student's name.}
\label{fig:biasesResults}
\end{figure}

Ultimately, the grade parameters presented in Table \ref{table:origin_sex_grade_seeds} have led to the distributions shown in Figure \ref{fig:biasesResults}, which are implicitly present in the MERIT Dataset. These grade distributions result from trying to replicate a representative demographic context for the United States (A) and Spain (B) given the exposed in \ref{annex:gradeBiases}.

\section{Experiments}
\label{sec:experiments}
We train one of the most relevant model families for VrDU tasks: the LayoutLM models. Specifically, we train LayoutLMv2 \cite{Xu2020a}, LayoutLMv3 \cite{Huang2022}, and LayoutXLM \cite{Xu2021} on the Token Classification task, which is the primary niche of the MERIT Dataset. The samples used to train LayoutLMv2 and v3 are in English, while those for training LayoutXLM are in Spanish. This demonstrates the multilingual versatility of both our dataset and our generation pipeline.

Given that the English and Spanish subsets include seven different school templates, we use samples from 5 schools to train and validate the model, reserving the remaining samples from the other two schools for testing. We decide to include only samples with Model A and Model B layouts as testing subsets (Figure \ref{fig:layoutFeatures}). To avoid excessively challenging the models when working with Blender-modified samples, we removed those samples with excessive folds and parts of the page with words outside the image margins. In addition, we work under the perfect OCR hypothesis, i.e., when testing, the model receives the original words and bounding boxes from the dataset, so no OCR induces downstream errors. We present the benchmark results in Table \ref{table:trainingResults}.

\begin{table}[ht]
\centering
\scriptsize
\begin{tabular}{lllllll}
\hline
& \textbf{Scenario 1} & \textbf{Scenario 2} & \textbf{Scenario 3} & FUNSD/ & &\\
& Dig./Dig. & Dig./Mod. & Mod./Mod & XFUND & &\\
\hline
&  {F1}&  {F1}&  {F1}&  {F1}&  {Lang.} &  {(Tr./Val./Test)} \\ \hline
LayoutLMv2 & 0.5536 & 0.3764 & 0.4984 & 0.8276 & Eng. & 7324/ 1831/ 4349\\
LayoutLMv3 & 0.3452 & 0.2681 & 0.6370 & 0.9029 & Eng. & 7324/ 1831/ 4349 \\
LayoutXLM & 0.5977 & 0.3295 & 0.4489 & 0.7550  & Spa. & 8115/ 2028/ 4426\\
\hline
\end{tabular}
\caption{LayoutLM family Benchmark on the MERIT Dataset. In scenario 1 (S1), the models are trained, validated, and tested using digital samples. In scenario 2 (S2), the models are trained and validated with digital samples but tested with a subset of Blender-modified samples. In scenario 3 (S3), the models are trained, validated, and tested using Blender-modified samples. The F1 scores refer to the base-case models. The FUNSD dataset is used for models in English, while the Spanish subset of XFUND is used for the non-English analysis.}
\label{table:trainingResults}
\end{table}

\section{Conclusions and contributions}
\label{sec:discussion}

\subsection{Conclusions} 
\label{subsec:Conclusions}
To analyze the training results with our dataset, we need to work from a perspective different from that of a paper presenting a model. We are presenting a dataset; therefore, we want this dataset to be relevant and pose a challenge for state-of-the-art models. In other words, we aim for reasonable results demonstrating our data's validity while showing room for improvement in the models. The results shown in Table \ref{table:trainingResults} can be analyzed in two ways:

First, we compare the results obtained by the LayoutLMv2/v3 and LayoutXLM models on the FUNSD/XFUND datasets with the results obtained using our dataset. The results on FUNSD/XFUND are noticeably better, indicating that our dataset presents a significant challenge for these models. Our analysis is based on the following points:

\begin{itemize}
\item The MERIT Dataset contains up to two orders of magnitude more labels (over 400 vs. 4), which presents a more demanding scenario for the models and results in lower metrics.
\item The MERIT Dataset has an order of magnitude more training samples and up to two orders of magnitude more test samples (for each language). On the one hand, the larger number of training samples provides the model with more examples to extract representative information and patterns. On the other hand, a larger test dataset is more comprehensive and challenging compared to FUNSD/XFUND (which both have only around 50 test samples per language).
\end{itemize}

Secondly, we compare how the models perform as the difficulty of the scenarios increases (horizontal axis in Table \ref{table:trainingResults}). We establish a baseline for each model where the models are trained with digital samples, which have minimal visual and layout noise. Once the baseline is established, we continue training with purely digital samples but test with samples more representative of real-world scenarios: Blender-modified samples that mimic actual conditions. In this case, the models face a sim-to-real gap, and as expected, the results deteriorate. Finally, we observe that when training with samples closer to real-world conditions, the gap between training and test distributions narrows, leading to improved performance compared to the previous scenario.

It is worth noting that we have identified some paradoxes. For example, it is surprising that LayoutLMv3 achieves better results in scenario 3 (with modified samples in both training and testing) compared to scenario 1 (with digital samples in both training and testing). Additionally, the analysis of the vertical axis in Table \ref{fig:biasesResults} reveals that LayoutLMv3, which is more powerful than LayoutLMv2 and LayoutXLM according to the authors' benchmarks, only shows improvement in scenario 3. However, a detailed analysis of the reasons behind this behavior is beyond the scope of this paper and will be addressed in future research.

\subsection{Main Contributions}
\label{subsec:mainContributions}
This publication contributes in two significant ways:

\begin{itemize}
    \item Dataset: We introduce the MERIT dataset: a multimodal, photorealistic, and exhaustively labeled dataset featuring image, text, and layout modalities in English and Spanish. It is up to 33k samples, including templates and graphic assets from 14 schools. This dataset stands out for several reasons:

    \begin{itemize}
        \item Synthetic Nature: It allows unrestricted use in any model without data protection concerns, a unique feature as no comparable dataset exists free from such restrictions. Additionally, its marginal generation cost is highly competitive \footnote{Samples were generated on an MSI Meg Infinite X 10SF-666EU with an Intel Core i9-10900KF and an Nvidia RTX 2080 GPU, running on Ubuntu 20.04}, approximately 2 seconds to generate one digital sample plus 34 seconds to modify it in Blender. This is significantly less than traditional human-labeling methods, which estimate the process on one hour per human-labeled document \cite{Xu2022}. Regarding energy consumption, the dataset is again highly competitive: we consumed 0.016 kWh/1000 samples when generating the digital samples and 0.366 kWh/1000 samples when modifying them in Blender, significantly less than text-to-image generative models with a median energy consumption of 1.35 kWh per 1000 unlabelled AI-generated images \cite{luccioni2023power}.

        \item Realism: We ensured the dataset reflects reality across all modalities:

        \begin{itemize}
            \item Photorealism: The dataset includes digital and physical document samples, with the former indistinguishable from actual digital documents and the latter featuring realistic scanning scenarios, including imperfect framing and paper imperfections.
            \item Layouts and Text: Inspired by real samples and utilizing a text editor for input, avoiding the common visual incongruence in AI-generated images with text. This approach ensures accuracy and consistency between the visual elements and labels.
        \end{itemize}
        \item Detailed Labeling: Our comprehensive and precise labeling approach eliminates the variability typically associated with human labeling efforts. By introducing a finer level of granularity, we significantly enhance the categorization of labels, elevating manually labeled datasets to a new order of magnitude in terms of the number of categories.
        \item Challenge Level: This dataset sets a new benchmark standard by creating an extensive range of label categories and providing numerous templates featuring diverse layouts, texts, and visual features. It presents a significantly more challenging benchmark compared to preceding datasets.
    \end{itemize}
    \item Pipeline: We also release the code for generating these samples, aiming for transparency and community contribution. This enables others to create samples tailored to specific needs, contributing expert knowledge, especially in crafting samples in non-Latin alphabets like Japanese, Russian, or Arabic.
    
    The generation pipeline offers extensive control and variability through:

    \begin{itemize}
        \item Templates: A user-friendly interface for custom layout and template design.
        \item Keywords: Automatic replacement and sampling of key information like student names and grades, allowing for customizability and variability.
        \item Graphic Assets: Users can personalize and position graphic elements like signatures and stamps, enhancing the dataset's realism and applicability.
    \end{itemize}
\end{itemize}

\subsection{Future Developments}
\label{subsec:futureDevelopments}
The dataset's current state facilitates a wide array of experiments with LLMs, leading the authors to exploit its present features for research in two key areas:

\begin{itemize}
    \item Benchmarking LLMs for specific tasks such as key information retrieval or VrDU, focusing on model families like LayoutLM. The objective is to contrast niche models against the broader goal of a Generalist AI. These studies will concentrate on factors like the energy consumption of models and the precision trade-off in tasks not yet mastered by generalist LLMs.
    
    \item Benchmarking LLMs in the context of bias detection. Many institutions, including offices, universities, and industries, increasingly depend on text-processing models like ChatGPT. These models are also employed in sensitive areas affecting individuals, such as job recruitment and university admissions. Despite their widespread use, there is a lack of public data on the ethical performance of these models. The MERIT Dataset addresses this gap, offering a substantial volume of intentionally biased and labeled data for multimodal models, encompassing image, text, and layout components. The research on this topic will study whether the suspicion of bias is real and the main forces driving biases.

\end{itemize}

Furthermore, the dataset's future development includes several initiatives:

\begin{itemize}
    \item Contextual improvements. When introducing biases into the dataset, it proves valuable to group subjects into knowledge blocks (natural sciences, social sciences, pure sciences, etc.). This way, it is possible to replicate behaviors that have traditionally been detected in PISA reports \cite{PISA2022}, where historically different patterns for men and women arise when dealing with different knowledge blocks.
    
    \item Broadening the Scope. While key information retrieval from school records poses substantial technical challenges (a vast array of classes and diverse layouts) or sensitivity to biases (e.g., grade biases linked to different personal conditions), the same technique for creating multimodal samples is adaptable to other significant domains. These include medical record analysis or the evaluation of official documents for social assistance processes.

    \item Enhancing Versatility. Current limitations in the generation pipeline appear when dealing with exhaustive characteristics of real-world documents. These include managing conditional information, like adding recovery grades if a student fails a subject or handling grades for a single subject spread over two semesters.
    
    \item Advancing Photorealism. 

\end{itemize}

\section*{Declaration of AI-assisted technologies in the writing process}
The authors used tools such as Grammarly and ChatGPT to correct spelling and enhance clarity, grammar, and sentence structure during the manuscript preparation. Subsequently, the authors thoroughly reviewed and edited the content
as required, assuming full responsibility for the publication’s content.

\section*{Data availability}
\label{sec:dataAvailability}
The MERIT Dataset and its generation pipeline are available on \href{https://huggingface.co/datasets/de-Rodrigo/merit}{Hugging Face} \footnote{Dataset on \href{https://huggingface.co/datasets/de-Rodrigo/merit}{Hugging Face}: https://huggingface.co/datasets/de-Rodrigo/merit} and \href{https://github.com/nachoDRT/MERIT-Dataset}{GitHub} \footnote{Code on \href{https://github.com/nachoDRT/MERIT-Dataset}{GitHub}: https://github.com/nachoDRT/MERIT-Dataset} respectively.

The training sessions from where we extract the metrics to discuss in Section \ref{subsec:Conclusions} are available on \href{https://wandb.ai/iderodrigo/MERIT-Dataset}{WandB}. \footnote{Training sessions on \href{https://wandb.ai/iderodrigo/MERIT-Dataset}{WandB}: https://wandb.ai/iderodrigo/MERIT-Dataset}. 

\section*{Acknowledgments}
\label{sec:acknowledgments}
The authors of this publication would like to thank the Chair for Smart Industry for providing the necessary resources to produce this research. In addition, they also acknowledge the contributions of Mauro Liz Soto as a research student. Lastly, the authors extend their gratitude to the Secretary's Office staff at Universidad Pontificia Comillas for granting access to authentic transcripts of records under the condition of a Non-Disclosure Agreement, enabling the creation of a dataset as closely aligned with reality as possible.

\appendix

\section{Grade biases}
\label{annex:gradeBiases}
The biases introduced into the MERIT Dataset stem from two sources: the origin of the name and the gender associated with the name. Table \ref{table:origin_sex_grade_seeds} indicates the parameters to model these biases. The authors have derived these parameters from the most objective data: PISA reports from various editions (with a consistent difficulty criterion for different international contexts, same test fields, application across regions of interest, etc.). The source for most of the data in Table \ref{table:PISA} comes from the 2022 PISA report \cite{PISA2022} (a period relevant to the data collection for training most modern LLMs). The data in these tables were obtained from the averages of the three exam sections: mathematics, reading comprehension, and science. The data for India (relevant for our USA school samples) was extracted from the 2009+ PISA report \cite{OCDEPISA2009} (since India has not participated in these reports since that date). On the other hand, the relevant data for Senegal (relevant for the Spanish schools) was obtained from the 2017 PISA for Development report \cite{SENEGAL2017}.

\begin{figure}
\centering
\includegraphics[width=1\columnwidth]{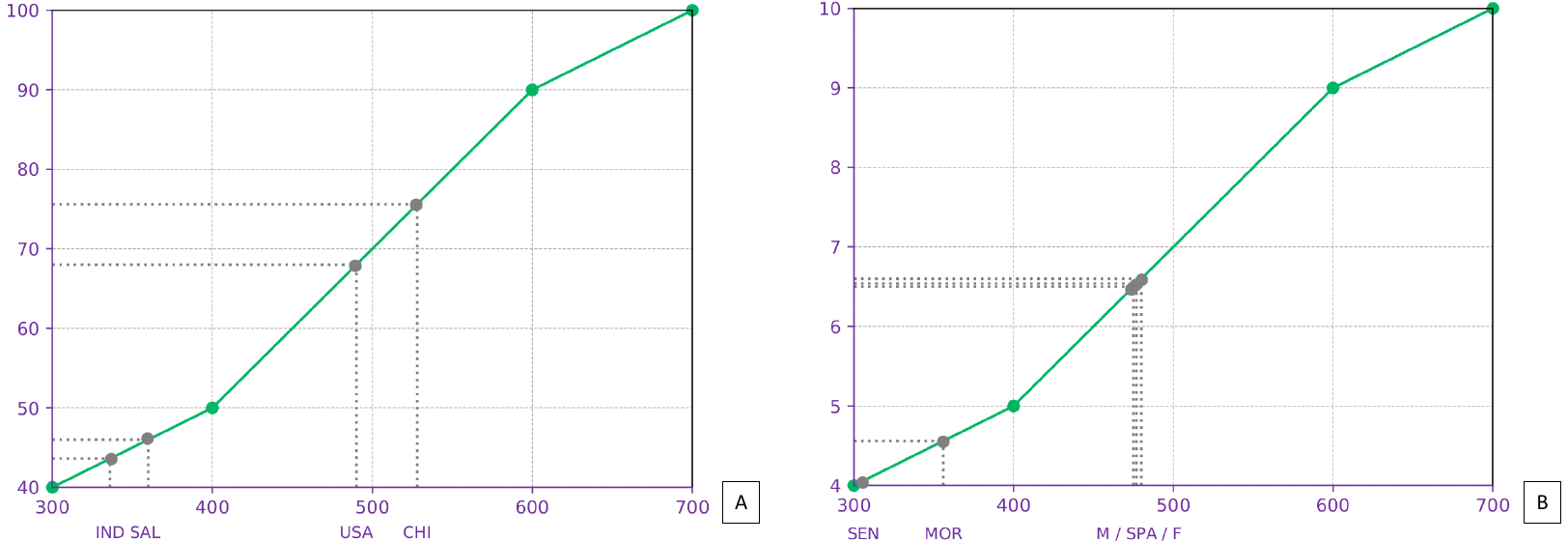}
\caption{PISA Scores translation to USA (A) and Spain (B) educative system scores used in Table \ref{table:origin_sex_grade_seeds}.}
\label{fig:grades}
\end{figure}

The \textit{PISA Score} to \textit{Educative System Score} (USA or Spain) equivalences outlined in Table \ref{table:PISA} were computed based on the criteria established in Figure \ref{fig:grades}, which in turn is estimated based on the clarifications made by PISA in their section \textit{Results: What do the test scores mean?} within their FAQs \cite{FAQsPISA}.

\begin{table}[htbp]
\centering
\small

\begin{tabular}{lllll}
\hline
\multicolumn{5}{c}{\textbf{Data from relevant countries}}         \\ \hline
Educative System & Origin & Country & PISA Score & System Score\\\hline
USA Schools & English & USA & 490 & 68\\
& Spanish & El Salvador & 360 & 46\\
& Chinese & China & 528 \tablefootnote{Available data for China only includes data from Macao and Hong Kong.} & 76\\
& Indian & India & 336 \tablefootnote{Data extracted from the PISA 2009+ report \cite{OCDEPISA2009}. Data from India only covers two regions whose scores appear separately. The score provided here is the average for those two regions.} & 44\\
\hline
Spanish Schools & Spanish & Spain & 477 & 6.54 \\
& Arabic & Morocco & 356 & 4.56\\
& Subsaharian & Senegal & 306 \tablefootnote{Data extracted from the PISA for Development report \cite{SENEGAL2017}.} & 4.06\\
\hline
\multicolumn{5}{c}{\textbf{Data in OCD Countries}}         \\ \hline
Gender &  &  & PISA Score & System Score\\ 
& & & & \tiny{(USA/Spain)} \\ \hline
Female & & & 480 & 66/6.60\\
Male & & & 475 & 65/6.50\\
\hline
\end{tabular}
\caption{PISA average scores for countries relevant to the demographic context in USA and Spain, the two considered academic systems in the MERIT Dataset.}
\label{table:PISA}
\end{table}


\bibliographystyle{elsarticle-num-names} 
\bibliography{OOL_bibliography}





\end{document}